\documentclass[letterpaper, 10 pt, conference]{ieeeconf}  

\IEEEoverridecommandlockouts                              

\usepackage{graphics} 
\usepackage{epsfig} 
\usepackage{mathptmx} 
\usepackage{times} 
\usepackage{amsmath} 
\usepackage{amssymb}  
\usepackage{todonotes}
\usepackage{etoolbox}

\usepackage[font=small,labelfont=bf]{caption} 

\usepackage{float}
\usepackage{hyperref}
\usepackage{sidecap}
\usepackage{lipsum} 

\title{\LARGE \bf
kPAM 2.0: Feedback Control for Category-Level Robotic Manipulation
}

\author{Wei Gao, Russ Tedrake 

\thanks{CSAIL, Massachusetts Institute of Technology, 77 Massachusetts Ave, Cambridge, USA. Emails: \{weigao, russt\}@mit.edu.}
}

\begin{document}

\makeatletter
\g@addto@macro\@maketitle{
  \begin{figure}[H]
  \setlength{\linewidth}{\textwidth}
  \setlength{\hsize}{\textwidth}
  \centering
  \includegraphics[width=\textwidth]{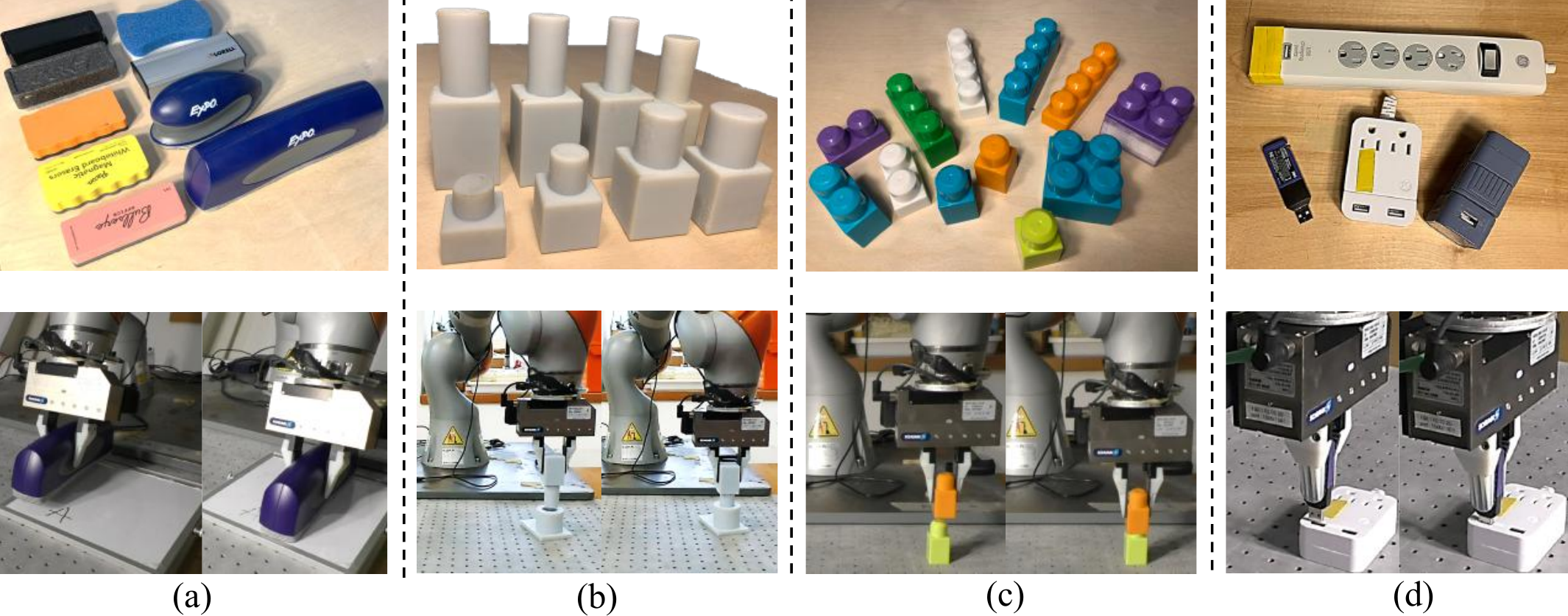}
  \caption{We explore generalizable, perception-to-action robotic manipulation for contact-rich tasks that can automatically handle a category of objects, despite large intra-category shape variation. Here we demonstrate: (a) wiping a whiteboard using erasers with different shape and size; (b)-(d) Peg-hole insertion for (b) 0.2 [mm] tight tolerance pegs and holes, (c) LEGO blocks and (d) USB ports. The particular novelty of these demonstrations is that our method automatically handles objects under significant intra-category shape variation (top row) without any instance-wise tuning, for each task (a)-(d). The video is on \href{https://sites.google.com/view/kpam2/}{\textcolor{blue}{\underline{https://sites.google.com/view/kpam2/}}}.
  \label{fig:experiments}
  }
  \end{figure}
}
\makeatother

\maketitle
\thispagestyle{empty}
\pagestyle{plain}

\begin{abstract}

In this paper we explore generalizable, perception-to-action robotic manipulation for precise, contact-rich tasks. In particular, we contribute a framework for closed-loop robotic manipulation that automatically handles a category of objects, despite potentially unseen object instances and significant intra-category variations in shape, size and appearance.
Previous approaches typically build a feedback loop on top of a realtime 6-DOF pose estimator.
However, representing an object with a parameterized transformation from a fixed geometric template does not capture large intra-category shape variation.
Hence we adopt the keypoint-based object representation proposed in kPAM\footnote{kPAM~\cite{manuelligao2019kpam} stands for \textbf{K}ey\textbf{P}oint \textbf{A}ffordance-based \textbf{M}anipulation.} \cite{manuelligao2019kpam} for category-level pick-and-place, and extend it to closed-loop manipulation policies with contact-rich tasks. We first augment keypoints with local orientation information. Using the oriented keypoints, we propose a novel object-centric action representation in terms of regulating the linear/angular velocity or force/torque of these oriented keypoints.
This formulation is surprisingly versatile -- we demonstrate that it can accomplish contact-rich manipulation tasks that require precision and dexterity for a category of objects with different shapes, sizes and appearances, such as peg-hole insertion for pegs and holes with significant shape variation and tight clearance. 
With the proposed object and action representation, our framework is also agnostic to the robot grasp pose and initial object configuration, making it flexible for integration and deployment.
The video demonstration and source code are available on \href{https://sites.google.com/view/generalizable-feedback/home}{\textcolor{blue}{\underline{this link}}}.

\end{abstract}

\section{Introduction}
\label{sec:intro}

Human can perform precise, reactive and dexterous manipulation while easily adapting their manipulation skill to new objects and environments. This remains challenging for robots despite obvious importance to both industrial and assistive applications. In this paper, we take a step towards this goal with emphasis on adaptability: the closed-loop, perception-to-action manipulation policy should generalize to a category of objects, with potentially unknown instances and large intra-category shape variations. Furthermore, the policy should be able to handle different initial object configurations and robot grasp poses for practical applicability.

While many works address robot grasping of arbitrary objects~\cite{ten2017grasp, schwarz2018fast}, these methods are typically customized to picking up the objects; extending them to other tasks is not straightforward. Contributions on visuomotor policy learning exploit neural network policies trained with data-driven algorithms~\cite{levine2016end, van2016stable, zhu2018reinforcement}, and many interesting manipulation behaviours emerge from them. 
However, how to efficiently generalize the trained policy to different objects, camera positions, object initial configurations and/or robot grasp poses remains an active research problem.

On the other hand, several vision-based closed-loop manipulation pipelines~\cite{kappler2018real, trikit, nvidiarbt} use 6-DOF pose as the object representation. They build a feedback loop on top of a real-time pose estimator. However, as detailed in Sec. 4 of kPAM~\cite{manuelligao2019kpam}, representing an object with a parameterized pose defined on a fixed geometric template, as these works do, may not adequately capture large intra-class shape variations. Thus, kPAM~\cite{manuelligao2019kpam} uses 3D keypoints as the object representation instead of 6-DOF pose for pick-and-place tasks. kPAM assumes an arbitrary rigid transformation can be applied to the object. This assumption is not true for contact-rich tasks in this work: although peg-hole insertion is eventually a rigid transformation of the peg, it is not easy to ``apply'' this rigid transformation.

\textbf{Contribution.} We propose (i) the first manipulation framework we know that is capable of precise, contact-rich manipulation while automatically generalizing to a category of objects. To achieve this, we adopt the keypoint-based object representation in kPAM~\cite{manuelligao2019kpam} and (ii) augment keypoints with the local orientation information. Using the oriented keypoint, (iii) we propose a novel object-centric action representation as the linear/angular velocity or force/torque of oriented keypoints. The object and action representations enable closed-loop policies and contact-rich tasks, despite significant intra-category shape variations of manipulated objects. (iv) We further show the proposed object and action representation lead to various additional benefits, which includes: the re-targeting to new grasp poses/initial object configurations, the processing of force/torque measurement for a category of objects, and significant simplification of kPAM category-level pick-and-place. (v) We experimentally validate our framework on a hardware robot with several challenging, industrially important tasks. Our demo is the first closed-loop, perception-to-action manipulation we know that can handle such a diversity of objects and task setups automatically.

Another desirable property of our framework is the extendibility. As shown in Sec.~\ref{subsec:general_formulation}, our framework includes a perception module and a feedback agent, establishes their interfaces but leaves the room for their actual implementation.
Thus, various existing model-based or data-driven algorithms for perception and control can potentially be plugged into our framework and automatically generalize to new objects and task setups, as long as the proposed object and action representation are used as their input/output.

This paper is organized as follows. Sec.~\ref{sec:related} reviews related works. Sec.~\ref{sec:formulation} describes our formulation of the generalized manipulation framework. Sec.~\ref{sec:pick_and_place} shows the significant simplification of the kPAM~\cite{manuelligao2019kpam} category-level pick-and-place.
Sec.~\ref{sec:results} presents hardware experiments on several challenging tasks, specifically showing generalization of our method. 

\section{Related Works}
\label{sec:related}

\subsection{Object Representation for Closed-Loop Manipulation}

Perhaps 6-DOF pose is the most widely used object representation for manipulation, thus pose estimation is studied extensively. Many datasets~\cite{wang2019normalized, xiang2014beyond} are annotated with pre-aligned templates, and pose estimators~\cite{wang2019normalized, sahin2018category} trained on them can produce a category-level estimation. Several teams~\cite{kappler2018real, trikit, nvidiarbt} incorporate realtime pose estimators into closed-loop manipulation pipelines and show impressive demos. To generalize these pipelines to a category of objects, a straightforward approach is combining them with category-level pose estimators. However, as shown in Sec. 4 of kPAM \cite{manuelligao2019kpam}, pose estimation can be ambiguous under large intra-category shape variations; a valid 6-DOF trajectory for one object can lead to physical infeasibility for other instances. A comparison is made in Sec.~\ref{subsec:failure_mode}.

On the other hand, many works train a visuomotor policy using data-driven algorithms~\cite{levine2016end, kumar2016optimal, van2016stable, zhu2018reinforcement}. The object representation (visual feature) in these methods is an internal state of the neural network. 
Some works exploited techniques such as autoencoders~\cite{levine2016end, van2016stable}, domain-randomization~\cite{andrychowicz2020learning, finn2016deep} and human demonstrations~\cite{zhu2018reinforcement, levine2016end}.
%
Compared with them, the key advantage of our framework is the automatic generalization to new object instances, camera positions, object initial configurations and robot grasp poses. On the other hand, many of these algorithms can potentially be integrated into our framework for category-level generalization and that would be a promising future direction.

\subsection{Robotic Manipulation with Proprioceptive Feedback}

There have been impressive works~\cite{tang2015learning, amanhoud2019dynamical, raibert1981hybrid} on robot control about industrially important tasks such as peg-hole insertion and polishing. By using joint torque sensors and/or 6-DOF force/torque sensors along with other proprioceptive sensors, the robot can perform impressive tasks, for instance peg-hole insertion with tight tolerance~\cite{tang2015learning} or polishing an non-flat surface with smooth trajectories~\cite{amanhoud2019dynamical}.
However, these methods typically assume perfectly known geometry with objects (tools) pre-fixed to robots. In this way, inaccuracy caused by visual perception and grasping is eliminated. 
For many tasks these prerequisites can be hard to satisfy.

\subsection{Manipulation at a Category Level}

Grasping algorithms \cite{ten2017grasp, schwarz2018fast} enable robots picking up arbitrary objects, and many of them have achieved impressive generality. Furthermore, several works~\cite{manuelligao2019kpam, qin2019keto, zakka2019form2fit} study pick \textit{and} place at a category level. kPAM~\cite{manuelligao2019kpam} proposed to use 3D keypoints as object representation. KETO \cite{qin2019keto} extends kPAM with self-supervised keypoint learning. Form2Fit~\cite{zakka2019form2fit} uses shape descriptors for object placement in assembly.

In this paper, we focus on generalizable manipulation with closed-loop feedback for contact-rich scenarios. Using an open-loop policy, as these previous works~\cite{manuelligao2019kpam, zakka2019form2fit, qin2019keto} do, typically cannot accomplish these tasks. A comparison is made in the Sec.~\ref{subsec:failure_mode}.

\begin{figure}[t]
\centering
\includegraphics[width=0.48\textwidth]{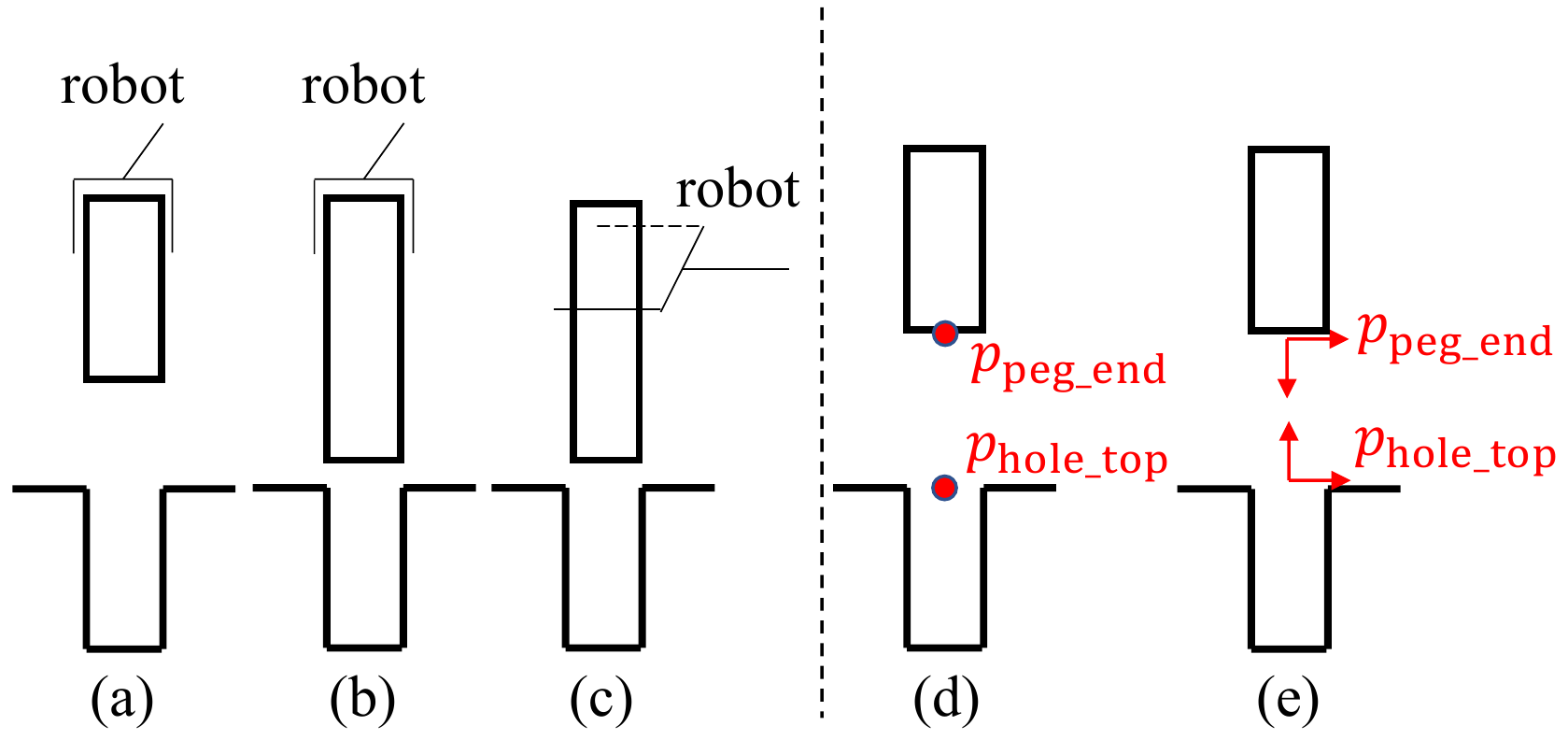}
\caption{\label{fig:oriented_keypoint} 
\textbf{The object representation using peg-hole insertion as an example}, as shown in (a). We would like the manipulation policy generalize to (b) a different peg; and (c) a different robot grasp pose. We adopt the semantic 3D keypoint proposed in kPAM~\cite{manuelligao2019kpam} as a local but task-specific object representation, as shown in (d). Since the task depends on the local relative orientation between the peg and hole, we augment keypoints with orientation information, as if a rigid coordinate is attached to each keypoint shown in (e).
}
\end{figure}

\section{Manipulation Framework}
\label{sec:formulation}

In this section, we discuss our formulation of the generalizable manipulation framework. Sec.~\ref{subsec:example} describes the approach using a concrete example, and Sec.~\ref{subsec:general_formulation} presents the general formulation. The subsequent sections discuss the details and extensions of the general formulation.

\subsection{Concrete Motivating Example}
\label{subsec:example}

Consider the task of peg-hole insertion, as illustrated in Fig.~\ref{fig:oriented_keypoint} (a). We want to come up with a manipulation policy that automatically generalizes to a different peg in Fig.~\ref{fig:oriented_keypoint} (b), and a different robot grasp pose in Fig.~\ref{fig:oriented_keypoint} (c).

kPAM~\cite{manuelligao2019kpam} proposed to represent the object by a set of semantic 3D keypoints. The motivation is: keypoints are well defined within a category while 6-DOF pose cannot capture large shape variation (see Sec. 4 of \cite{manuelligao2019kpam} for details). We adopt this idea and choose two keypoints: the $p_{\text{peg\_end}}$ that is attached to the peg and the $p_{\text{hole\_top}}$ that is attached to the hole, as shown in Fig.~\ref{fig:oriented_keypoint} (d). Similar to kPAM, we assume that we have a keypoint detector, for instance a deep network, that can produce these specified keypoints in real-time.

These two keypoints provide the location information. However, the peg-hole insertion task also depends on the relative orientation of the peg and hole. Thus, we augment keypoints with orientation information, as if a rigid coordinate is attached to each keypoint, as shown in Fig.~\ref{fig:oriented_keypoint} (e). For the peg-hole insertion task, we let the $z$ axis of the $p_{\text{peg\_end}}$, $p_{\text{hole\_top}}$ be the axis of the peg and hole, respectively. The $x$ axis of $p_{\text{peg\_end}}$, $p_{\text{hole\_top}}$ can be chosen arbitrarily, but when the $x$ axes of $p_{\text{peg\_end}}$ and $p_{\text{hole\_top}}$ are aligned the peg should be able to insert into the hole.

\begin{figure}[t]
\centering
\includegraphics[width=0.4\textwidth]{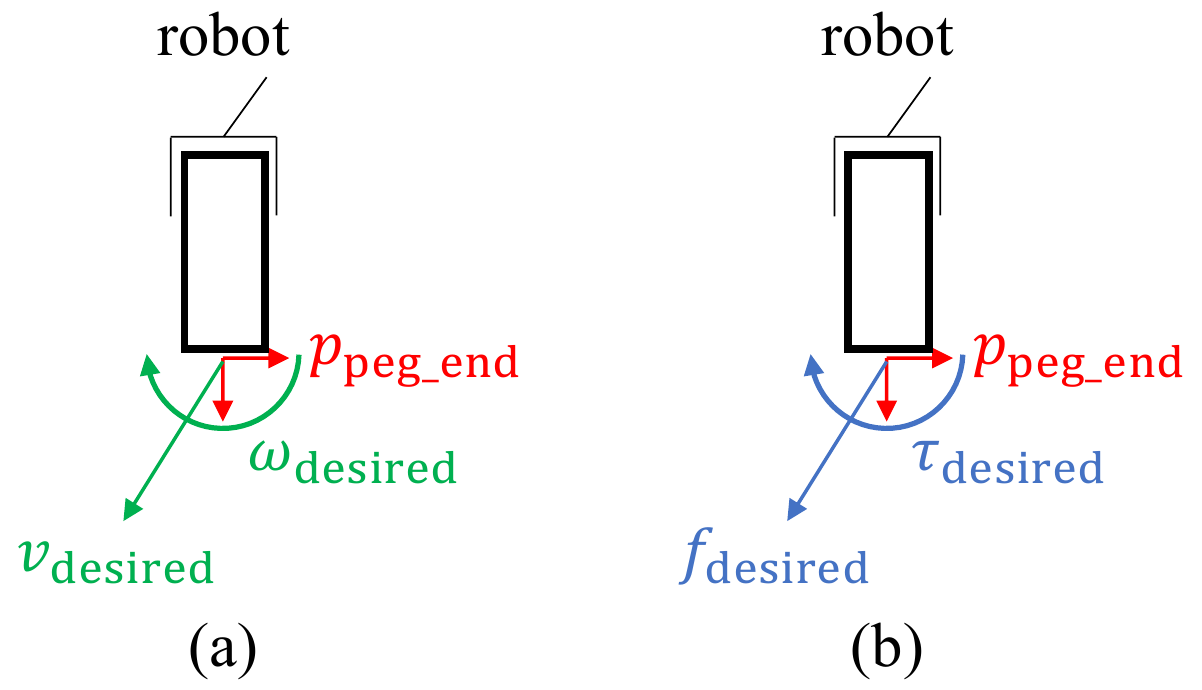}
\caption{\label{fig:action_space} 
\textbf{Overview of the object-centric action representation}. With the oriented keypoint in Fig.~\ref{fig:oriented_keypoint} (e) as the object representation, the action can be represented as: (a) the desired linear/angular velocity of an oriented keypoint; or (b) the desired force/torque of an oriented keypoint. Note that these two action representations are agnostic to the robot grasp pose and unrelated geometric details (see Sec.~\ref{subsec:example}).
}
\end{figure}

The coordinate in Fig.~\ref{fig:oriented_keypoint} (e) is also used to illustrate 6-DOF pose in the literature. The key difference between the oriented keypoint and 6-DOF pose is: the oriented keypoint is a \textit{local} but task-specific characterization of the object geometry, while pose with geometric template is global. The choice of a local object representation is inspired by the observation that in many manipulation tasks, only a local object part interacts with the environment and is important for the task. For instance, the $p_{\text{peg\_end}}$ keypoint only characterizes a local object part that will be engaged with the hole, and it does not imply task-irrelevant geometric details such as the handle grasped by the robot. This locality enables generalization to novel objects as the unrelated geometric details are ignored. A more detailed discussion is in Sec.~\ref{sec:discussion}. 

\begin{figure*}[t]
\centering
\includegraphics[width=0.9\textwidth]{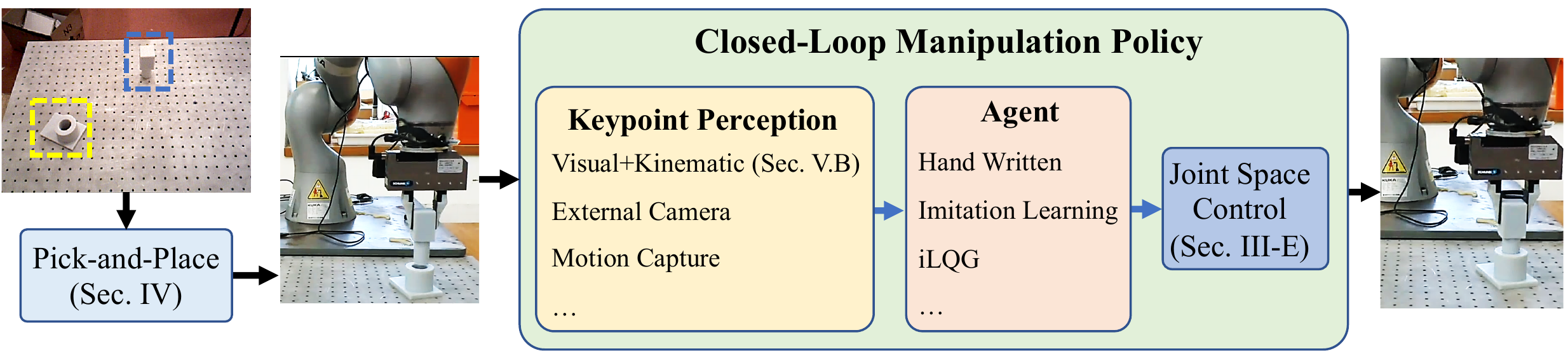}
\caption{\label{fig:pipeline} 
\textbf{Overview of the manipulation framework}. The closed-loop policy consists of 1) a perception module that produces oriented keypoints in real time; 2) an agent with the state and action space shown in Fig.~\ref{fig:oriented_keypoint} and Fig.~\ref{fig:action_space}, respectively; 3) a joint-space controller that maps agent outputs to joint-space commands. Note that many different implementations of the perception module and agent can be used within our framework and the resulting pipeline automatically generalize to new objects and task setups. For many applications the objects are randomly placed initially. In this scenario, we perform a kinematic pick-and-place to move the object to some desired initial condition (for instance moving the peg right above the hole), from where the closed-loop policy starts operating (see Sec.~\ref{sec:pick_and_place} for details). 
} 
\end{figure*}

As illustrated in Fig.~\ref{fig:action_space}, with the oriented keypoint as the object representation we propose to represent the robot action as either: 1) the desired linear and angular velocity of the $p_{\text{peg\_end}}$ keypoint in Fig.~\ref{fig:action_space} (a); or 2) the desired force/torque at the $p_{\text{peg\_end}}$ keypoint in Fig.~\ref{fig:action_space} (b). 
%
Importantly, these two action representations are defined w.r.t only a local part of the object. Because these actions are not defined w.r.t the robot, our method is agnostic to the robot grasp pose. Similarly, our method can handle a variety of objects as actions are also agnostic to irrelevant geometric details (such as the handle grasped by the robot).
These actions can be mapped to joint space commands, as described in Sec.~\ref{subsec:joint_space}.


Suppose we have implemented an agent (which can be a model-based controller or a neural network policy) using the object and action representations mentioned above as the input and output, together with a perception module that produces the required keypoints in real-time and a joint-level controller that maps the agent output to joint command, then the resulting manipulation policy would automatically generalize to different objects and robot grasp poses, for instance the ones in Fig.~\ref{fig:oriented_keypoint} (a), (b) and (c).
Even if the policy doesn't directly transfer due to unmodeled factors, it would be a good initialization for many data-driven or model-based algorithms~\cite{kumar2016optimal, levine2016end, schulman2017proximal}.

\subsection{General Formulation}
\label{subsec:general_formulation}

We can think of a robot as a programmable force/motion generator~\cite{holladay2019force}.
We propose to represent the task-specific motion profile as the motion of a set of oriented keypoints, and the force profile as the force/torque w.r.t some keypoints.

Thus, given a category-level manipulation problem we propose to solve it in the following manner. 
First the modeler selects a set of 3D oriented keypoints that capture the task-specific force/motion profile. Once we have chosen keypoints, the manipulation framework can be factored into: 1) the perception module that outputs oriented keypoints from sensory inputs; 2) the agent that takes the perceived keypoints as input and produces the desired linear/angular velocity or force/torque of an oriented keypoint as output; 3) the joint-space controller that maps the agent output to joint-space command. An illustration is shown in Fig.~\ref{fig:pipeline}. The framework can be extended with force/torque measurements and the generalization to different object initial configurations, as shown in Sec.~\ref{subsec:ft_measurement} and Sec.~\ref{subsec:generalize_initial}. 
For many applications, objects are randomly placed initially. In this case, we perform a kinematic pick-and-place to move the object to some desired initial configurations, from where the closed-loop policy starts, as shown in Sec.~\ref{sec:pick_and_place}. To make the overall manipulation operation generalizable, all these components should work for a category of objects.

It should be emphasized that our framework establishes the interfaces (input/output) of the perception module and closed-loop agent, but is agnostic to their implementation.
The only requirement is that for the perception module it should output oriented keypoints in real time, and for the agent it should use the state and action space mentioned above.
There are many solutions for both of them. 
For instance, in our experiment we combine the wrist-mounted camera and robot kinematics for keypoint perception (see Sec.~\ref{subsec:percep_impl}).
Alternatively, external cameras or motion capture markers can also be used for the keypoint tracking. 
Similarly one might explore various model-based or data-driven controllers as the feedback agent according to the task in hand.
In particular, many data-driven agents~\cite{kumar2016optimal, schulman2017proximal, levine2016contact} are agnostic to the state (object) and action representation, thus can be used directly without modification.
Integrating these perception module and controllers into our framework would achieve automatic generalization to new objects and task setups, as long as the proposed object and action representation are used as the input/output.

\begin{figure}[t]
\centering
\includegraphics[width=0.4\textwidth]{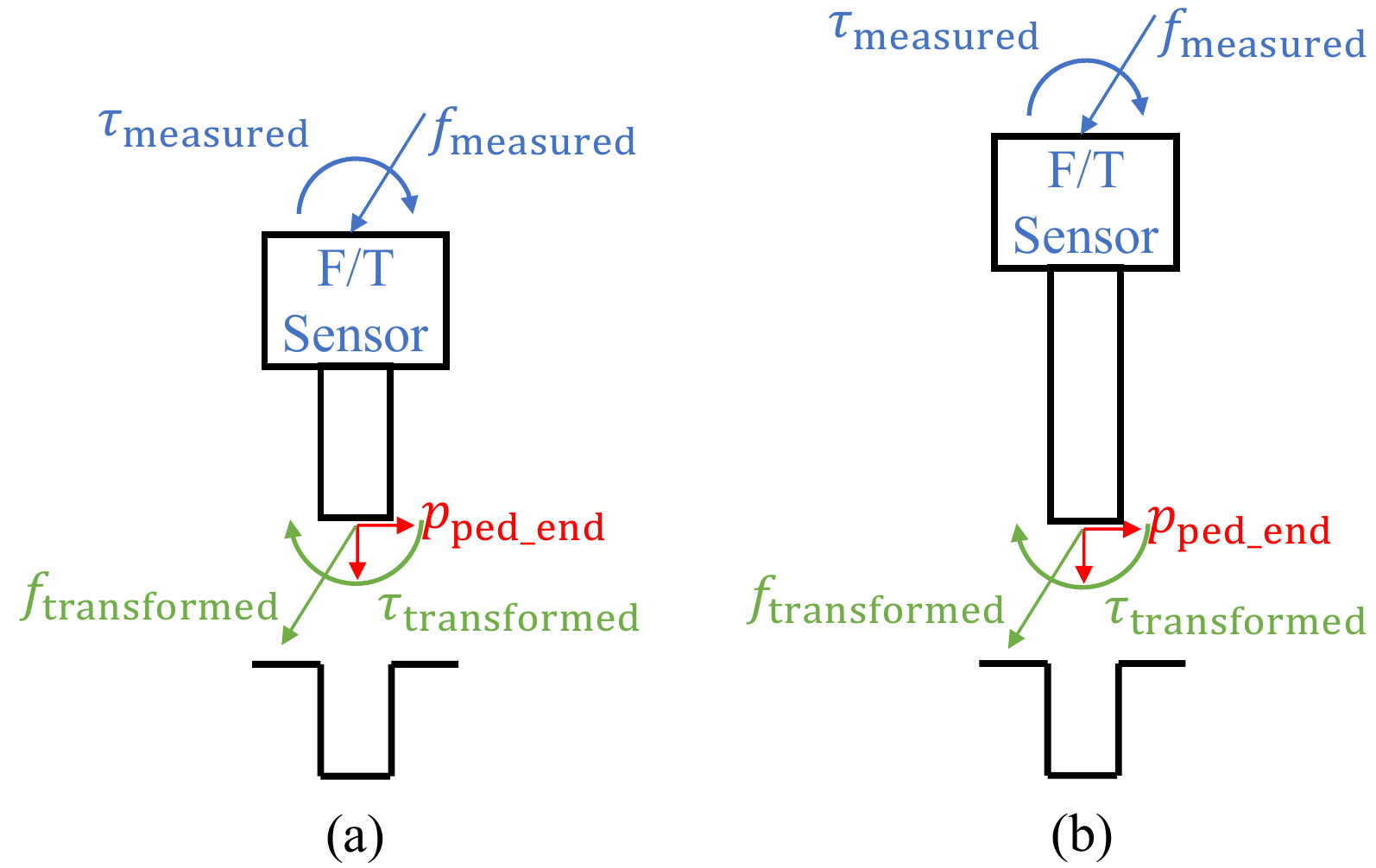}.
\caption{\label{fig:ft_measurement} 
\textbf{The processing of force/torque measurement in our framework}. For a wrist-mounted force/torque sensor, its raw measurements $f_\text{measured}$ and $\tau_\text{measured}$ are not invariant to the object geometry and grasp pose. Thus, we propose to transform them to an oriented keypoint ($p_\text{peg\_end}$ here), as the transformed measurement captures the task-specific force/torque profile. The closed-loop agent takes the transformed force/torque measurement as input would generalize w.r.t object geometry.
}
\end{figure}

\subsection{Force/Torque Measurement}
\label{subsec:ft_measurement}

Some robots are equipped with wrist-mounted force/torque sensors or joint torque sensors. For contact-rich manipulation tasks, it’s very beneficial to use this information as the input of the agent. However, the raw output from these sensors varies with the object geometry and robot grasp pose, as shown in Fig.~\ref{fig:ft_measurement}. As a result, directly feeding these measurements into the agent does not generalize automatically.

The solution to this problem is to transform the measured force/torque to the kinematic frame of an oriented keypoint, as shown in Fig~\ref{fig:ft_measurement}. Using the peg-hole insertion as an example, we can transform the force/torque measurement from the robot wrist to the coordinate of $p_{\text{peg\_end}}$, as if a “virtual sensor” is mounted at $p_{\text{peg\_end}}$. 
Comparing with the raw measurement from a sensor fixed w.r.t the robot, using a virtual sensor makes the measurement independent of unrelated object geometry and robot grasp pose.

If the robot is equipped with joint torque sensors, we can also estimate the force/torque by assuming the robot has no other contact. Let $J_{\text{peg\_end}}$ be the Jacobian that maps robot joint velocity to the linear/angular velocity of $p_{\text{peg\_end}}$, the force/torque $f_\text{estimated} \in R^{6}$ can be estimated as
\begin{equation}
f_\text{estimated} = \text{argmin}_f |J_{\text{peg\_end}}^T f - \tau_\text{external}|^2
\end{equation}
\noindent where $\tau_\text{external}$ is the measured external joint torque. Here we assume the gravity torque has been compensated.

\vspace{-0.3em}
\subsection{Generalization w.r.t Global Rigid Transformation}
\label{subsec:generalize_initial}

Suppose we want to re-target the peg-hole insertion policy to a hole at a different location.
Intuitively, this re-targeting is essentially a rigid transformation of the manipulation policy. Can we somehow ``apply" this transformation directly?

In our framework, both the agent input (oriented keypoints and force/torque w.r.t keypoints) and output (linear/angular velocity or force/torque of an oriented keypoint) are expressed in 3D space. In other words, we can apply a rigid transformation to both the agent input and output. This property provides generalization w.r.t the global rigid transformation. Before feeding the input to the agent, we can transform the input from the world frame to some ``nominal frame”. After agent computation, we can transform its output back to the world frame. The ``nominal frame” can be chosen arbitrary, for instance in the peg-hole insertion task we can align it with the initial configuration of $p_{\text{hole\_top}}$. Thus, the global rigid transformation is transparent to the agent.

On the contrary, many existing works~\cite{levine2016end, zhu2018reinforcement, florence2019self} use input/output in joint space or image space (raw image or 2D keypoints), on which a rigid transformation cannot be applied.
Thus, re-targeting agents in these methods to new initial configurations and camera poses may need re-training.

\subsection{Joint Space Control}
\label{subsec:joint_space}

The agent output the desired linear/angular velocity or force/torque of an oriented keypoint $p$. An important observation is: if we assume the object is rigid and grasp is tight (grasped object is static w.r.t the gripper), then the object can be regarded as part of the robot and consequently standard joint-space controllers can be used map these agent outputs to joint-space commands. This generalizes the ``object attachment" in collision-free manipulation planning~\cite{diankov2010automated}.

With this observation, we discuss several possible joint-space controller according to robot interfaces. Let $J_p$ be the Jacobian that maps joint velocity $\Dot{q}$ to linear/angular velocity of $p$. A straightforward method to transform the commanded velocity $v_p$ into joint velocity command $\Dot{q}_\text{desired}$ is
\begin{equation}
    \Dot{q}_\text{desired} = \text{argmin}_{\Dot{q}}|J_p \Dot{q} - v_p|^2 + \text{reg}(\Dot{q})
\end{equation}
\noindent where $\text{reg}(\Dot{q})$ is a regularizer term. If the robot driver accepts joint velocity commands, we can send $\Dot{q}_\text{desired}$ to the robot directly. Similarly, the desired force/torque $F_p$ can also be transformed into joint space by 
\begin{equation}
    \tau_\text{desired} = J_p^T F_p + g
\end{equation}
\noindent where $g$ is the gravitational force in joint space. Here we ignore the inertia and Coriolis force of the robot.

Since standard joint-space controllers can map the agent output to joint commands, more sophisticated controllers can also be used and might provide better tracking performance, for example the impedance controller in \cite{nakanishi2008operational}. The detailed discussion is omitted as they are out of our scope.

\begin{figure}[t]
\centering
\includegraphics[width=0.49\textwidth]{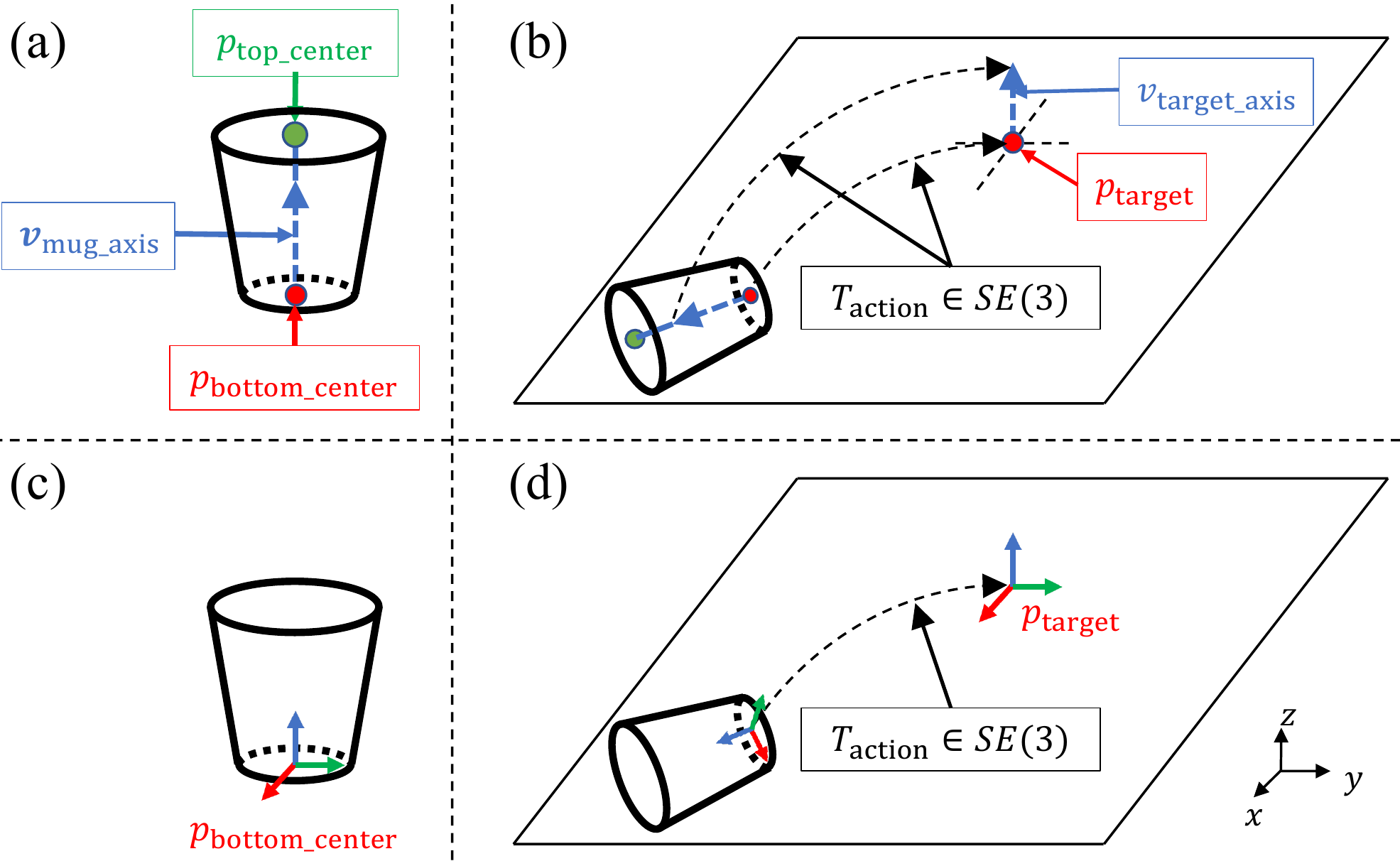}
\caption{\label{fig:kpam_review} 
\textbf{Overview of kPAM~\cite{manuelligao2019kpam} pipeline for category-level pick-and-place and its simplification with oriented keypoints}. kPAM is used as a pre-step of the closed-loop policy in our framework, as shown in Fig.~\ref{fig:pipeline}. Using the original mug demo of kPAM as an example: (a) In kPAM the object is represented by a set of semantic 3D keypoints. 
(b) The rigid transformation $T_\text{action}$, which represents the robot pick-and-place action, is solved to move $p_\text{bottom\_center}$ to the target location $p_\text{target}$ and align the mug axis with its target $v_\text{target\_axis}$. (c) In this paper we propose to add orientation information to the $p_\text{bottom\_center}$ keypoint. (d) The desired configuration of the mug can be encoded as the target configuration of $p_\text{bottom\_center}$, and $T_\text{action}$ is the relative transform between $p_\text{bottom\_center}$ and its target configuration. 
Note that both the original kPAM in (b) and the new formulation in (d) generalize to mugs with different shape and size.
More details are in Sec.~\ref{sec:pick_and_place}.
}
\end{figure}

\section{Pick-and-Place Manipulation}
\label{sec:pick_and_place}

In many applications, objects are randomly placed initially, potentially in a clutter.
For this scenario, we use a two-step manipulation scheme: the robot first perform a kinematic pick-and-place to singulate the object and move it to some desired initial configuration (for instance move the peg right above the hole), then the closed-loop policy described in Sec.~\ref{sec:formulation} starts from that initial configuration. We stress that even with this pick-and-place, the re-targeting to different initial configuration in Sec.~\ref{subsec:generalize_initial} is still necessary, as the closed-loop policy should generalize to different placement locations (the hole locations in our example). To make the entire manipulation operation generalizable, this pick-and-place step should also works for a category of objects.

kPAM~\cite{manuelligao2019kpam} is devoted to this task, as generalizable pick-and-place is important for many applications.
Here we show how the orientation information of keypoints can be used to significantly simplify the kPAM pipeline, as shown in Fig.~\ref{fig:kpam_review}.

\subsection{Preliminary: kPAM~\cite{manuelligao2019kpam} Pipeline for Pick-and-Place}
\label{subsec:kpam_review}

Here we briefly review the kPAM~\cite{manuelligao2019kpam} pipeline. In kPAM, each object is represented by a set of semantic 3D keypoints $p \in R^{3 \times N}$, where $N$ is the number of keypoints. Using the original mug manipulation demo in kPAM as an example, we can represent the mug by two keypoints: $p_\text{top\_center}$ and $p_\text{bottom\_center}$, as shown in Fig.~\ref{fig:kpam_review} (a). These keypoints are detected from raw sensory input such as RGBD image. 

In kPAM, the robot pick-and-place action is represented as a rigid transformation $T_\text{action} \in SE(3)$ on the object, 
and keypoints $p$ on that object would be transformed to $T_\text{action} p \in R^{3 \times N}$.
The object target configuration is defined as a set of geometric costs/constraints on the transformed keypoint $T_\text{action} p$. For example, to place the mug upright at some target location $p_\text{target}$, the planned robot action $T_\text{action}$ should satisfy
\begin{align}
    ||T_\text{action} p_\text{bottom\_center} - p_\text{target}|| &= 0 \\
    ||\text{rotation}(T_\text{action}) v_\text{mug\_axis} - (0,0,1)^T|| &= 0 \\
    \text{where: } v_\text{mug\_axis} = \text{normalized}(p_\text{top\_center}-&p_\text{bottom\_center})
\end{align}
\noindent Note that the costs/constraints encoding of the object target configuration remains valid for mugs with different shape and size. After solving the optimization problem and grasping the object, we can apply $T_\text{action}$ to the object, which is essentially a rigid transformation on the robot end effector.

\subsection{Simplification of kPAM using Oriented Keypoint}

As shown in Fig.~\ref{fig:kpam_review} (c), we can add orientation information to the $p_\text{bottom\_center}$ keypoint: the $z$ axis of $p_\text{bottom\_center}$ is aligned with $v_\text{mug\_axis}$ in Eq.~(6), while the $x$ axis of $p_\text{bottom\_center}$ is chosen randomly since the mug is symmetric. Then, the target configuration of the mug can be represented as a target configuration of $p_\text{bottom\_center}$, as shown in Fig.~\ref{fig:kpam_review} (d). The robot pick-and-place action $T_\text{action}$ in Sec.~\ref{subsec:kpam_review} is the relative transformation between $p_\text{bottom\_center}$ and its target configuration. Note that this formulation also generalizes to mugs with different shape, size and topology.

By adding the orientation information to the keypoint, in many applications (for example all the demos in kPAM~\cite{manuelligao2019kpam}) we can avoid setting up costs/constraints and solving an optimization problem to find $T_\text{action}$. This demonstrate the benefit of the oriented keypoint as a more informative local attention mechanism for robot manipulation.

\section{Results}
\label{sec:results}

We instantiate our framework on a hardware robot and demonstrate a variety of contact-rich manipulation tasks.
The particular novelty of these demonstrations is that our method handles objects with large intra-category shape variations without any instance-wise tuning. An overview of experiments is in Fig.~\ref{fig:experiments}, and the detailed setup is in the Appendix. The video is on \href{https://sites.google.com/view/kpam2/}{\textcolor{blue}{\underline{https://sites.google.com/view/kpam2/}}}.


\subsection{Task Description}
\label{subsec:task_descri}

\begin{figure}[t]
\centering
\includegraphics[width=0.25\textwidth]{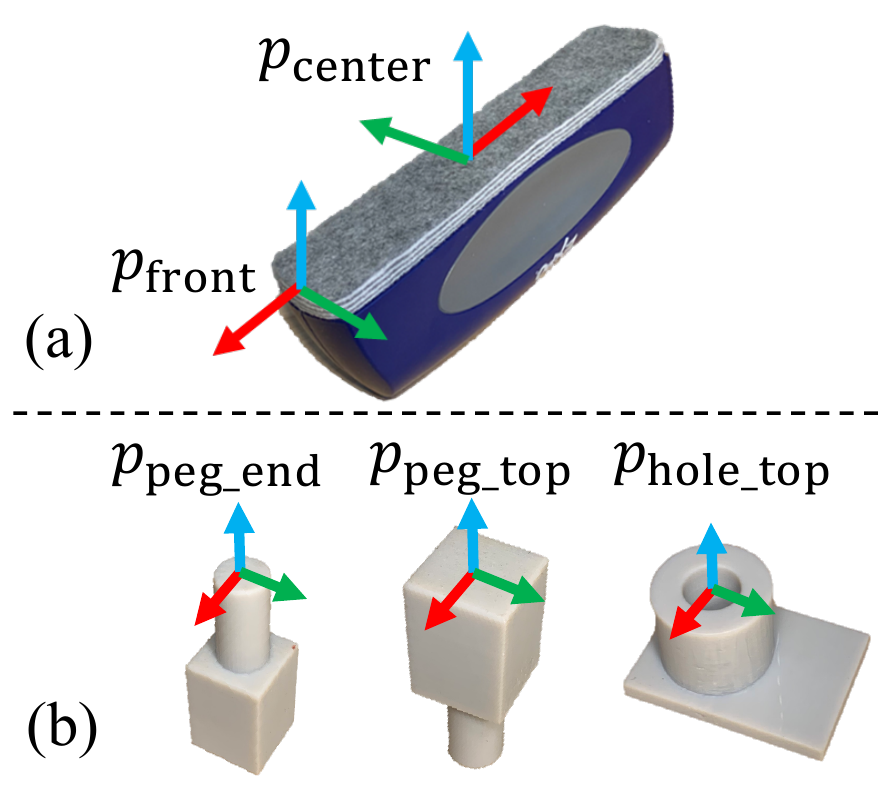}
\caption{\label{fig:specified_keypoint} 
\textbf{The specified keypoints used in our experiment}. (a) Two keypoints are detected for the whiteboard wiping task. (b) Two keypoints are detected for the peg, one keypoint is detected for the hole. For the manipulation of LEGO blocks in Fig.~\ref{fig:experiments} (c), the $p_\text{peg\_top}$ for another LEGO block is used as $p_\text{hole\_top}$. Please refer to Sec.~\ref{subsec:task_descri} for more details.
}
\end{figure}

\noindent \textbf{Whiteboard Wiping: } The robot must detect the whiteboard eraser, pick it up and use it to erase a small whiteboard, as shown in Fig.~\ref{fig:experiments} (a). We use two oriented keypoints for this task: $p_\text{front}$ and $p_\text{center}$ as shown in Fig.~\ref{fig:specified_keypoint} (a). For a successful wiping, the $x$-$y$ plane trajectory of $p_\text{front}$ should be aligned with the edge of the whiteboard, while the $z$ axis force on $p_\text{center}$ must be regulated to ensure the eraser is in contact with the whiteboard. We set the nominal $z$ axis force to be 10 [N] and implement the agent as a linear feedback controller to track the force on $z$ axis and position on other dimensions. 
The robot needs to deal with whiteboard erasers with significant shapes and sizes variation.

\vspace{0.3em}
\noindent \textbf{Peg-hole Insertion: } The robot must detect the peg and hole, pick the peg up and insert into the hole. We use three groups of objects: 1) 3D-printed pegs and holes with 0.2 [mm] clearance in Fig.~\ref{fig:experiments} (b); 2) LEGO blocks in Fig.~\ref{fig:experiments} (c); 3) USB drive and ports in Fig.~\ref{fig:experiments} (d). The same code is used for all three object groups. 
Due to the graspability limitation of the USB drive, we pre-fix it to the robot gripper, while the USB port as the ``hole" is detected from visual perception.

We generally follow the peg-hole insertion framework in Fig. 7 of \cite{tang2015learning} to implement the agent. Different from \cite{tang2015learning}, we use keypoints instead of the peg pose, the keypoint linear/angular velocity as the action representation instead of peg linear/angular velocity, and the transformed force/torque measurement (Sec.~\ref{subsec:ft_measurement}) instead of raw data, for category-level generalization. As lead-through demonstration in \cite{tang2015learning} is infeasible on our robot, we use a compliance controller (Sec. 2 in~\cite{morel1998impedance}) instead of the GMM regressor 
in Eq. (17) of~\cite{tang2015learning}.
Suggested by \cite{tang2015learning}, we use periodic switching between closed-loop and feedforward control (the direct and indirect control in \cite{tang2015learning}), as a random perturbation proven to be helpful in their experiment.
%

We also try to learn an agent by imitating successful trails, following almost exactly the setup of~\cite{tang2015learning}. The learned agent works on all three peg-hole categories (printed peg-hole, LEGO and USB), but it doesn't outperform the original controller. Later we found it is because the learned agent almost reproduces the original controller. We expect learning from human demonstration would be a promising future direction.

\begin{figure}[t]
\centering
\includegraphics[width=0.36\textwidth]{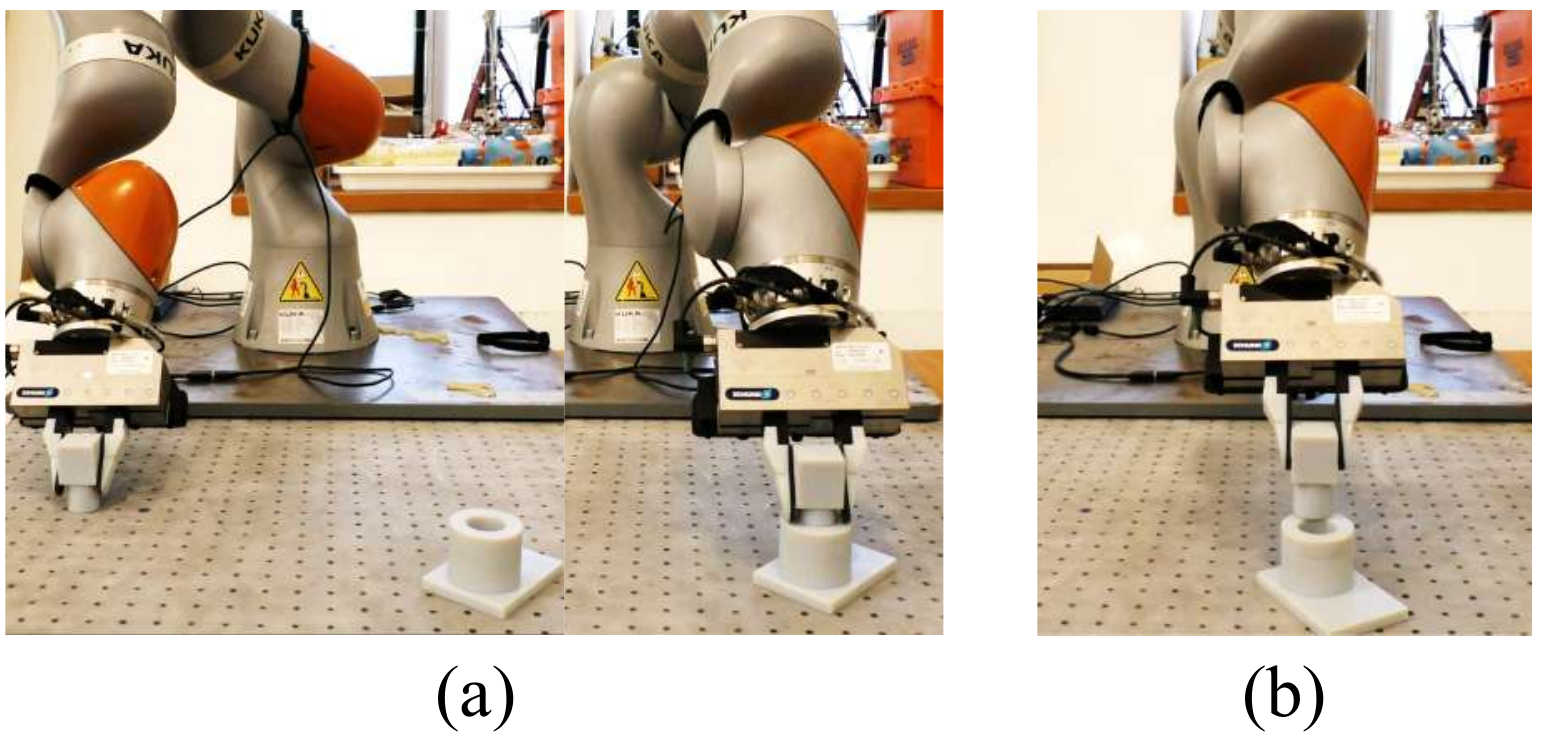}
\caption{\label{fig:failure_mode} 
\textbf{Typical failure modes}. (a) Grasp failure. (b) The keypoint perception error is too large such that the closed-loop agent can't correct it with feedback.
}
\end{figure}

\subsection{Perception Implementation}
\label{subsec:percep_impl}

We use robot kinematics to track oriented keypoints in real-time and use it as the input of the closed-loop agent.
Suppose we know oriented keypoints relative to the robot gripper, then when robot moves we can compute oriented keypoints in the world frame using the forward kinematics. 

We use a robot wrist-mounted camera to perform object detection, keypoint detection and grasp planning with the method in kPAM~\cite{manuelligao2019kpam}.
Given visual perception results, we execute robot grasping and compute the keypoints expressed in the robot gripper frame, using the keypoints in the world frame (from camera perception) and the robot gripper pose (from robot kinematics). After grasping, we use robot kinematics for real-time keypoint tracking and feed the result into the closed-loop agent, as mentioned above. Please refer to the Appendix for more perception implementation details.

We stress that our framework is not restricted to this perception implementation, although it is easy to realize and good enough for our experiment. It is an interesting direction to test other keypoint trackers, for instance ones based on external cameras or motion-tracking markers.

\subsection{Experimental Result}
\label{subsec:failure_mode}

The failure rates of our method are summarized in Table~\ref{table:success_rate}. For the wiping task, we mark a trial as failure if 1) the discrepancy between $p_\text{front}$ with the whiteboard edge is larger than 2 [cm]; or 2) the $z$ axis force on $p_\text{center}$ is less than 5 [N]. For the peg-hole insertion task, we mark a trial as a failure if the peg is not inserted into the hole. Our method is first compared with an open-loop baseline similiar to kPAM~\cite{manuelligao2019kpam}, Form2Fit~\cite{zakka2019form2fit} and KETO~\cite{qin2019keto}: for wiping task the open-loop policy replays the $p_\text{front}$ trajectory, for peg-hole insertion the open-loop policy always commands a downward motion. 
Our method has a much lower failure rate, as shown in Table.~\ref{table:success_rate}.

For the wiping task, it is crucial to measure and regulate the contact force in a closed-loop manner else the eraser would not touch the whiteboard. Thus, an open-loop policy typically cannot successfully erase texts on the whiteboard.

For the peg-hole insertion task, the typical accuracy of the visual keypoint detection is about 5 [mm] when the distance from the camera to objects is about 80 [cm]. The perception error is much larger than the clearance (the clearance is 0.2 [mm] for printed pegs and holes, almost zero for LEGO blocks and USB ports), which requires the agent to correct itself using closed-loop feedback with the measured keypoints and force/torque. For the printed pegs and holes, the agent can tolerate some perception error (please refer to appendix for more details), thus the failure rate is decent and much lower than the open-loop kinematic policy. However, if the perception error is too large the feedback agent wouldn't be able to correct it, as shown in Fig.~\ref{fig:failure_mode} (b). The LEGO blocks have large chamfers, which makes the insertion much easier. 

On the other hand, the USB port is much more demanding on the perception accuracy (roughly 3 [mm] error on the shorter side of USB would result in failure). We use a two-step perception scheme: the first coarse step roughly locates the object; then we move the wrist-mount camera closer to the object and perform the second, more accurate perception. 
In this way, we can reduce the perception error to 2 [mm].



\begin{table}[t]
    \centering
  \caption{\#Failure/\#Trial (Failure Rate) Comparison}
  \label{table:success_rate}
 \includegraphics[width=\linewidth]{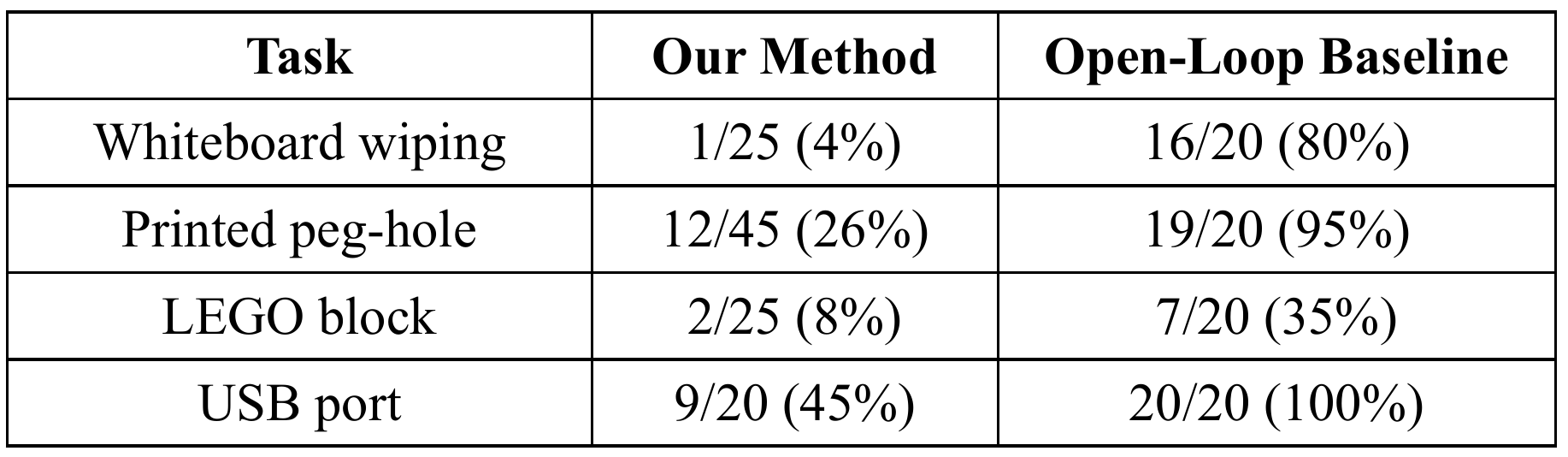}
\end{table}

\begin{table}[t]
  \caption{Summary for 6DOF-Pose Baseline}
  \label{table:pose_failure}
 \includegraphics[width=\linewidth]{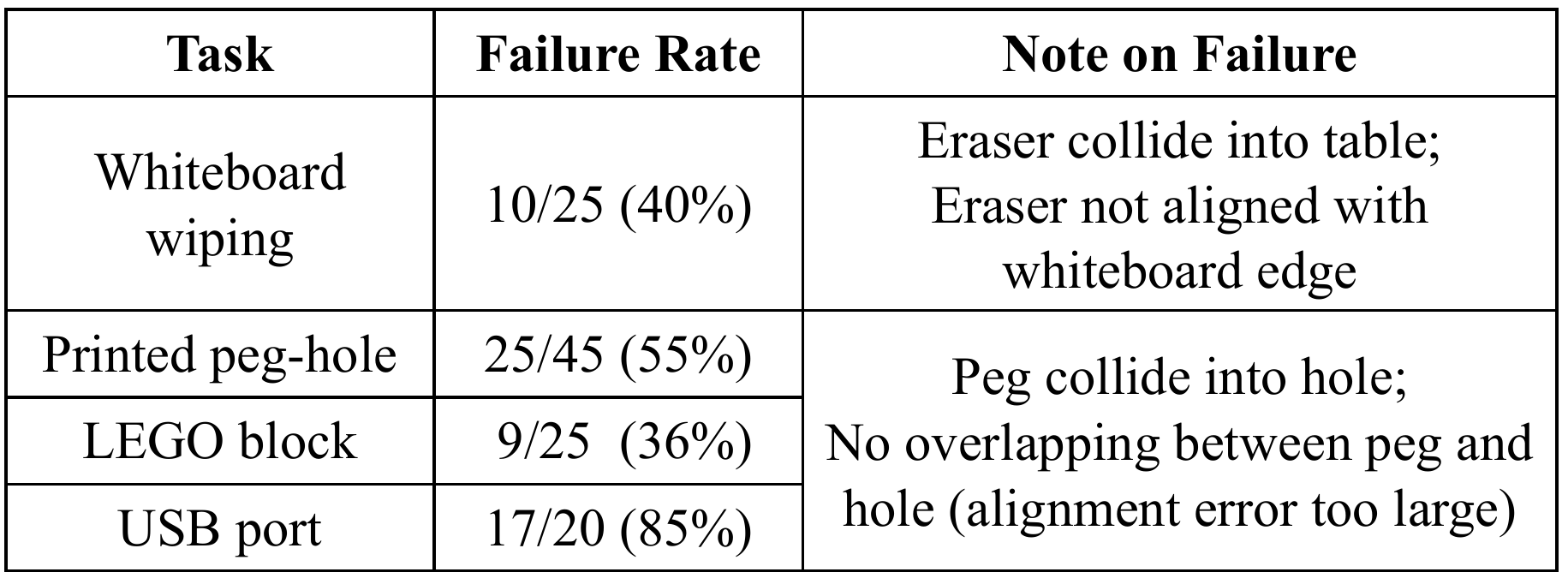}
\end{table}

To demonstrate the superiority of our keypoint formulation over pose-based methods, we implement a pose estimator and test it on our setups. The pose estimator is the same as the baseline (Fig. 4) in kPAM~\cite{gao2019kpam}: first initialize the alignment with detected keypoints, then perform ICP fitting between the observed point cloud and geometric template to get the 6-DOF pose. As demonstrated in kPAM (Fig. 5 of kPAM), a valid 6-DOF pose trajectory for one object can lead to physical infeasibility for another instance. For this reason, many trials would fail kinematically and some of them can be dangerous and require safety stop, as shown in Table~\ref{table:pose_failure}. Our approach has a much better performance.

\section{Discussion and Comparison}
\label{sec:discussion}

In this section we compare the oriented keypoint in Sec.~\ref{sec:formulation} with existing object representations for robotic manipulation.

As mentioned in Sec.~\ref{subsec:example}, the keypoint is a local but task-specific representation of the object. The motivation is that for many robot manipulation tasks only a local object part is important. Because of the locality, there might be multiple keypoints on an object for some tasks, for instance two keypoints are used in the whiteboard wiping task shown in Fig.~\ref{fig:specified_keypoint}. On the contrary, the mapping between the object and the 6-DOF pose is one-to-one.

On the other hand, 6-DOF pose is defined as a transformation on a template model. Pose estimators typically rely on the geometric matching between the observation (image, point cloud) and the template, where the matching is used either directly or as ground-truth for network training. Thus, 6-DOF pose with geometric template is a global and task-agnostic representation of the object geometry.
However for a category of objects with different instances, the geometric matching can be ambiguous, as shown in~\cite{manuelligao2019kpam}.

For manipulating one object with perfectly known geometry, the 6-DOF pose and keypoint (with or without orientation) are equivalent: pose can be estimated given detected keypoints; inversely we can get keypoints from pose, by annotating keypoints on the template and transforming those keypoints with an estimated pose. Thus, keypoints can contain more information than pose. Existing agents that succeed with 6-DOF pose on a specific object, 
should perform equally well or better with keypoints on that object.
Moreover, keypoints have the benefit of category-level generalization, making it a promising alternative over 6-DOF pose.


On the other hand, end-to-end methods~\cite{levine2016end, martin2019variable} typically use implicit object representations as internal states of the neural network. Compare with them, the oriented keypoint provides the prior knowledge about which part of the object is task-relevant. Our framework automatically handles new object geometry because task-irrelevant object parts are ignored. This prior knowledge is not available for a end-to-end method. They must discover it from the training data with many different objects, which can be expensive in practice.

Although 3D keypoints in~\cite{manuelligao2019kpam} are not oriented, the local orientation is implicitly used to encode the object target configuration in their pipeline, for instance the “put mugs on a table” task (Fig. 6). This highlights the importance of local orientation and the benefit to incorporate them explicitly.

\section{Limitation and Future Work}
\label{sec:limitations}

Our method built upon the sparse keypoint representation cannot handle tasks that require dense object geometry. For example if the peg is larger than the hole, then our framework cannot realize the task is impossible. Other tasks that require dense geometry include collision-free trajectory planning. To resolve it, a promising method is to exploit shape completion algorithm to obtain the dense geometry of objects.

We assume the object is static w.r.t the gripper. This essentially restricts the applied force on the object: the applied force should be balanced by contact forces that lie in the static friction cones. To mitigate this, the simplest method is to plan a stable, tight grasp. Another possible way is using external cameras for perception, instead of the wrist-mount camera in our experiment. In this way we can detect the relative motion between the gripper and object.

Benefit from the flexibility of our framework, it can be integrated with many other perception modules and feedback agents. For example, we might consider human demonstration instead of imitating a hand-written policy in our experiment, or explore other camera setups and neural network structures. We believe these are promising future directions to explore.

\section{Conclusion}
\label{sec:conclusion}

We present a novel framework for closed-loop, perception-to-action manipulation that handles a category of objects, despite large intra-category shape variation. To achieve this, we introduce the oriented keypoint as an object representation for manipulation, and purpose a novel action representation on top of that. 
Moreover, our framework is agnostic to the robot grasp pose and object initial configuration, makes it flexible for integration and deployment. Extensive hardware experiments demonstrate the effectiveness of our method.

\noindent {\footnotesize \textbf{Acknowledgments} This work was supported by  National Science Foundation Award no. EFMA-1830901, Amazon.com Services LLC and Lincoln Laboratory/Air Force PO\# 7000470769. The views expressed in this paper are those of the authors and not endorsed by the funding agencies.}



{\small
\bibliographystyle{abbrv}
\bibliography{paper.bib}
}

\end{document}